\documentclass[
]{ceurart}

\usepackage[caption=false]{subfig}
\usepackage[export]{adjustbox}
\usepackage{siunitx}
\usepackage{booktabs}
\usepackage{tabularx}
\usepackage{makecell}
\usepackage{colortbl}
\usepackage{multirow}
\usepackage{tablefootnote} 

\usepackage{xcolor}
\usepackage{color}

\usepackage{hyperref}
\usepackage{enumitem}
\usepackage{listings}
\usepackage{float}
\usepackage{listings}
\usepackage{enumitem}
\usepackage{orcidlink} 
\newcommand{\mypar}[1]{\vspace{0.5pt}\noindent\textbf{#1.}}

\newcommand{\toolname}{FairLoop}

\begin{document}
\copyrightyear{2025}
\copyrightclause{Copyright for this paper by its authors.
  Use permitted under Creative Commons License Attribution 4.0
  International (CC BY 4.0).}

\conference{Proceedings of the Best BPM Dissertation Award, Doctoral Consortium, and Demonstrations \& Resources Forum co-located with 23rd International Conference on Business Process Management (BPM 2025), Seville, Spain, August 31st to September 5th, 2025.}

\title{FairLoop: Software Support for Human-Centric Fairness in Predictive Business Process Monitoring}

\author[1]{Felix M\"ohrlein}[%
orcid=0009-0004-7142-6102,
email=fmoehrlein@hof-university.de
]
\cormark[1]
\cortext[1]{Corresponding author.}

\author[2]{Martin K\"appel}[%
orcid=0009-0003-3420-8037,
email=martin.kaeppel@fau.de
]

\author[3]{Julian Neuberger}[%
orcid=0009-0008-4244-7659,
email=julian.neuberger@uni-bayreuth.de
]

\author[2]{Sven Weinzierl}[
orcid=0000-0003-2268-7352,
email=sven.weinzierl@fau.de
]

\author[1]{Lars Ackermann}[%
email=lackermann@hof-university.de
]

\author[2]{Martin Matzner}[
orcid=0000-0001-5244-3928,
email=martin.matzner@fau.de
]

\author[3]{Stefan Jablonski}[%
email=stefan.jablonski@uni-bayreuth.de
]

\address[1]{University of Applied Science Hof, Alfons-Goppel-Platz 1, 95028 Hof, Germany}
\address[2]{Friedrich-Alexander-University Erlangen-Nuremberg, F\"urther Straße 248,
90429 Nuremberg, Germany}
\address[3]{University of Bayreuth, Universit\"atsstra{\ss}e 30, 95444 Bayreuth, Germany} 


\begin{abstract}
Sensitive attributes like gender or age can lead to unfair predictions in machine learning tasks such as predictive business process monitoring, particularly when used without considering context. We present \toolname{}\footnote{See \url{https://github.com/fmoehrlein/FairLoop} for code, video, and live demo.}, a tool for human-guided bias mitigation in neural network-based prediction models. \toolname{} distills decision trees from neural networks, allowing users to inspect and modify unfair decision logic, which is then used to fine-tune the original model towards fairer predictions. Compared to other approaches to fairness, \toolname{} enables context-aware bias removal through human involvement, addressing the influence of sensitive attributes selectively rather than excluding them uniformly.
\end{abstract}


\maketitle

\section{Introduction}
Predictive business process monitoring (PBPM)~\cite{maggi2014predictive} has emerged as an important subfield within business process management research~\cite{ackermann2024recent}. Unlike descriptive and diagnostic analysis of event data, which primarily focus on past events, predictive business process monitoring provides real-time decision support during process execution by forecasting the future evolution of process instances. This encompasses, among others, predictions of activities that will be executed next, remaining cycle times, and potential outcomes of the process~\cite{di2022predictive}. From a business standpoint, these predictions offer significant value, as early insights enable proactive interventions to avoid undesirable outcomes or facilitate the thorough preparation of upcoming steps~\cite{marquez2017predictive}. 

Since prediction models are increasingly applied in business processes with critical decisions, such as medical treatment processes, concerns about their fairness and the ethical implications of their use have gained increasing attention~\cite{de2024achieving}. In this context, fairness refers to the equitable treatment of individuals or groups~\cite{dwork2011fairnessawareness}, particularly when sensitive attributes like gender, age, race, or socioeconomic status are involved~\cite{Oneto2020}. These attributes often reflect historical inequalities or systemic biases that are embedded in the data~\cite{Barocas2016}. 

Many PBPM techniques use machine learning algorithms to automatically construct predictive models from even log data~\citep{weinzierl2024machine}. In doing so, the prediction models may inadvertently inherit embedded biases, resulting in undesired or discriminatory predictions. 

Existing approaches, therefore, often remove bias by removing sensitive attributes completely. However, this strategy is insufficient, as sensitive attributes can be crucial depending on their respective role in different activities within the process. For instance, while the gender of a patient should not influence appointment allocation, it may be essential when prescribing medication. Bias introduced by sensitive attributes that violates fairness is referred to as \emph{negative bias} (e.g., using gender in appointment allocation). In contrast, bias that is necessary for the model's intended purpose is termed \emph{positive bias} (e.g., considering gender for medication prescriptions). Therefore, a more nuanced approach is required, which restricts the usage of sensitive attributes only in contexts where they are harmful. 

This issue is aggravated by the fact that state-of-the-art prediction models usually rely on neural networks that inherently represent a black box, i.e., their decision-making process is not transparent~\cite{Kaeppel.2024}. As a result, it is often unclear to what extent sensitive attributes are being used unfairly for predictions. In~\cite{Kaeppel2025Fairness}, a novel approach is introduced to address these two challenges. First, a decision tree (that is inherently interpretable) is distilled from the black box prediction model in order to visualize the decision-making process. Domain experts can then identify potentially discriminatory decision rules and modify the decision tree to align with the desired fair behavior. Next, the adjusted decision tree is used to re-label the training dataset to reflect unbiased decisions. Finally, the re-labeled training dataset is used to fine-tune the prediction model in order to eliminate the bias. While technically feasible, the success of this approach largely depends on the expertise of domain experts and the decisions they make.

Therefore, it is essential to support domain experts with appropriate tools that allow the identification of the bias in prediction models and offer an easy adjustment of the distilled decision trees, especially in the case of complex ones. To this end, we present~\toolname{}\footnote{See \url{https://github.com/fmoehrlein/FairLoop} for code, video, and live demo.}, a tool designed to facilitate human-guided bias removal in prediction models. \toolname{} supports the described workflow of knowledge distillation, human-guided bias removal, and retraining.

\section{System Overview}

\begin{figure}[bt]
    \centering
    \includegraphics[width=\linewidth]{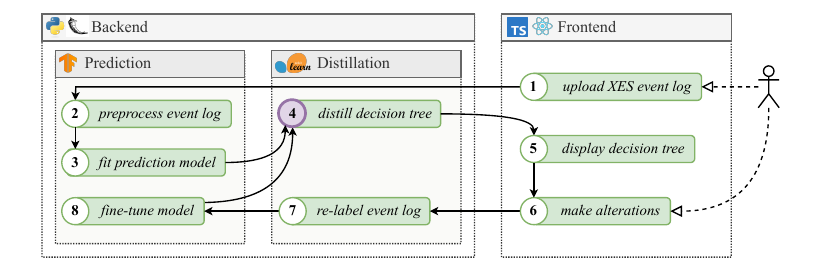}
    \vspace{-5mm}
    \caption{Overview of the modular architecture of \toolname{}.}
    \label{fig:system-architecture}
\end{figure}

We designed the architecture of our tool with extensibility in mind, splitting it into two separate modules, as shown in Fig.~\ref{fig:system-architecture}.

The \textit{\textbf{frontend}} allows the user to conveniently interact with the system via their browser.
It is written in TypeScript and React\footnote{See \url{https://www.typescriptlang.org/} and \url{https://react.dev/} respectively, both last accessed June 26, 2025.}, and guides the user through uploading an event log in the XES format\footnote{See \url{https://xes-standard.org/}, last accessed June 26, 2025.}, to the second component, i.e., the \textit{\textbf{backend}} module, which is written in Python 3.10 and utilizes Flask for providing a REST API\footnote{See \url{https://www.python.org/} and \url{https://flask.palletsprojects.com/en/stable/} respectively, both last accessed June 26, 2025}.
The backend is split into two sub-modules for \textit{\textbf{prediction}} and \textit{\textbf{distillation}}, which allows easy integration of new prediction or distillation approaches into our system in the future.
The prediction sub-module currently uses TensorFlow\footnote{See \url{https://www.tensorflow.org/}, last accessed June 26, 2025} for training a multi-layer perceptron (MLP) on event log data.
This MLP is then passed to the distillation sub-module, which creates a decision tree using the scikit-learn library\footnote{See \url{https://scikit-learn.org/stable/index.html}, last accessed June 26, 2025.}.
We build the decision tree from a dataset that consists of all possible prefixes in the event log and the corresponding prediction made by the MLP.
The decision tree is then converted into a custom implementation, which allows us to modify the structure of the decision tree after training.
This is done by sending the decision tree to the frontend, where the user can inspect decisions and remove those that are negatively biased (see Fig.~\ref{fig:decision-tree-screenshot}).
After any number of alterations, the user can trigger re-labeling the training data and fine-tuning the prediction model on the modified dataset.
To this end, we assign each prefix in the event log the target predicted by the (now fairer) decision tree, which in turn lets the MLP learn to make fairer predictions during fine-tuning.
Performance metrics such as accuracy, F1 score, precision, and recall for the different predictive models are shown for comparison, evaluated on the original (unmodified) dataset. As a result, performance may appear to decrease, since fairer predictions can conflict with the potentially biased ground truth in the original data.
After this step, the user can view the effects of their alterations by distilling a new decision tree from the fine-tuned MLP model.
This distill-alter-tune cycle can be repeated any number of times, allowing the user to iteratively guide the MLP model toward increasingly fair behavior, as outlined in Fig.~\ref{fig:system-architecture}.

\begin{figure}[bt]
    \centering
    \includegraphics[width=1\linewidth]{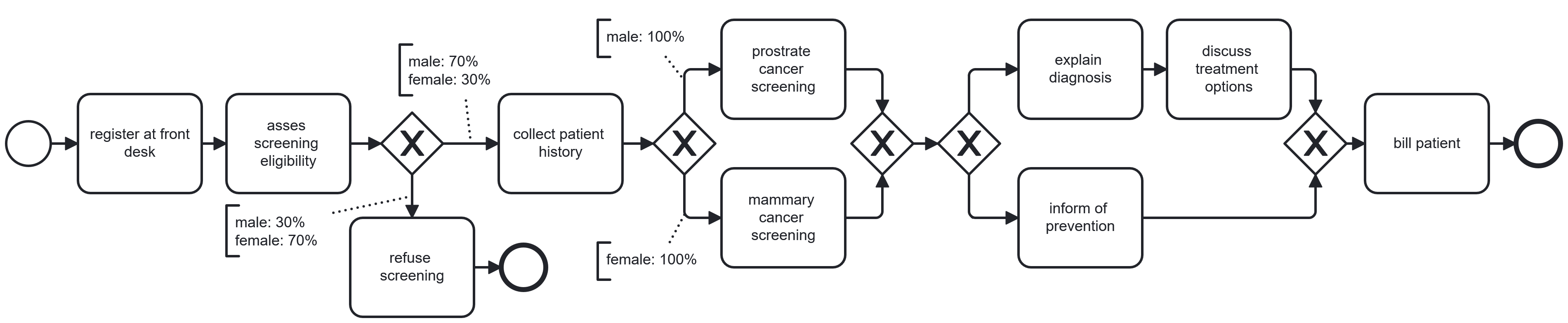}
    \caption{Diagram describing the underlying process of the simulated \textit{cancer screening} event log used for the decision tree in Fig.~\ref{fig:decision-tree-screenshot}. Decisions that are influenced by the attribute \textit{gender} are annotated with the respective transition probabilities. While deciding whether a patient should undergo \textit{prostate screening} or \textit{mammary screening} based on the attribute \textit{gender} can be a fair decision in the context of the process, refusing screening when \textit{gender = female} is a negatively biased decision.}
    \label{fig:cancer_screening}
\end{figure}

\begin{figure}[bt]
    \centering
    \includegraphics[width=0.8\linewidth]{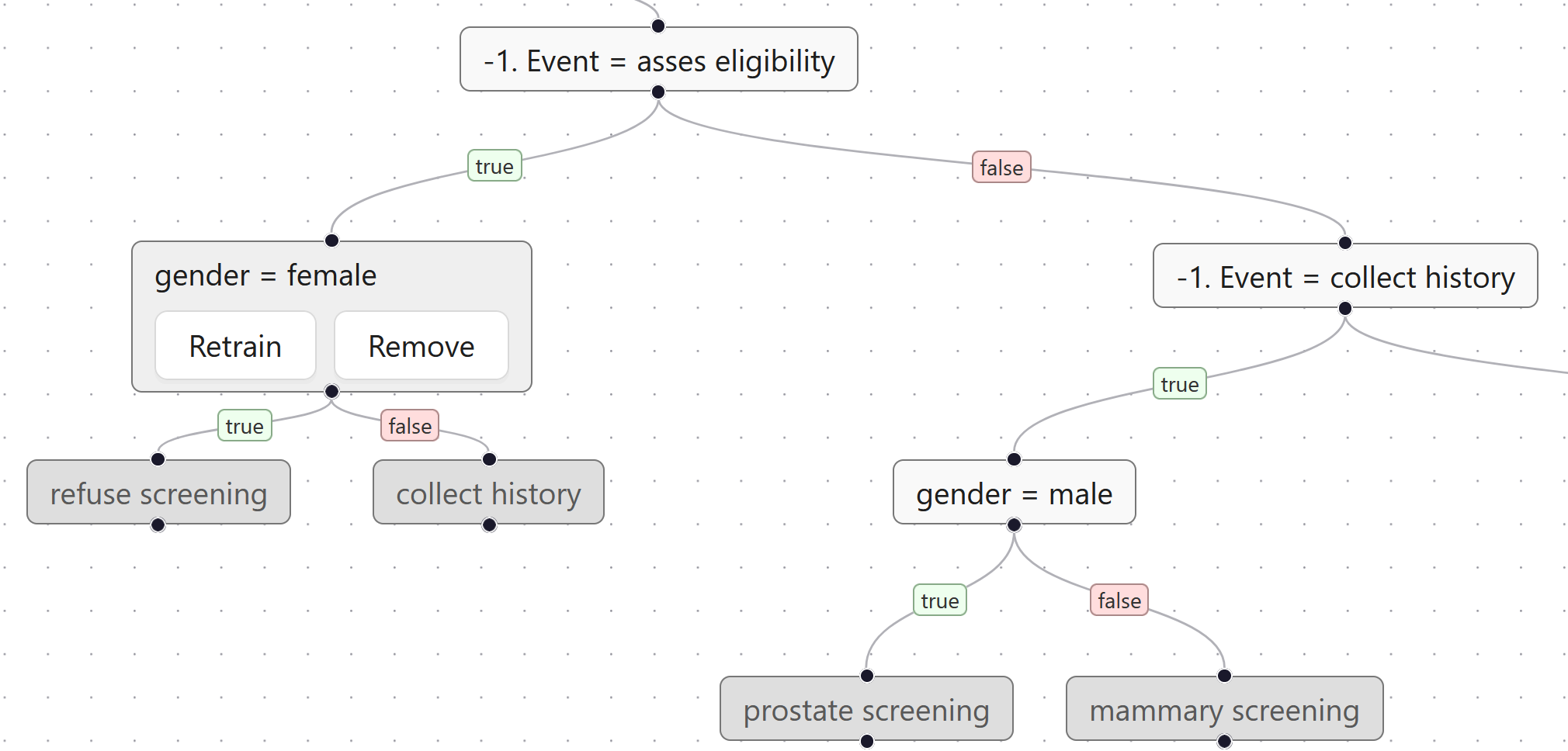}
    \caption{Screenshot of a portion of the decision tree displayed in the frontend,  
    for a simulated event log modeled after the \textit{cancer screening} process  
    depicted in Fig.~\ref{fig:cancer_screening}.  
    The node responsible for unfairly refusing screening when \textit{gender = female} is selected; it can be removed either directly or by retraining its subtree without using the \textit{gender} attribute.  
    Other nodes using \textit{gender} to introduce positive bias remain unaffected.} 
    \label{fig:decision-tree-screenshot}
\end{figure}

\section{Key Innovations}
\toolname{} makes MLPs used for PBPM explainable using a novel form of visualization: distilled decision trees. By translating the behavior of MLPs into these transparent rule-based representations, \toolname{} enables users to inspect and reason about predictions in a structured and intuitive way. This approach is not yet present in other graphical tools for PBPM, such as Nirdizati~\cite{nirdizati}, which also aim to provide interpretability of trained models.

Another innovation of \toolname{} lies in its approach to bias mitigation. Existing tools, such as the \textit{Discrimination-aware Decision Tree}~\cite{fairness_tree} plugin for ProM\footnote{See \url{https://promtools.org/}, last accessed June 26, 2025.}, an open source framework for process mining, represent early efforts to address fairness in PBPM. That plugin extends traditional decision trees by incorporating fairness constraints, aiming to reduce discrimination in predictions.
However, this approach comes with two major limitations. 
First, it is restricted to decision trees as the underlying PBPM model. In contrast, \toolname{} uses decision trees only as an intermediate representation to support user interaction. The actual predictions are made by more expressive MLPs, resulting in a better predictive performance. This allows \toolname{} to combine the interpretability of decision trees with the predictive strength of neural networks.
Second, the fairness constraints are applied uniformly, removing the influence of sensitive attributes, such as \textit{gender}, regardless of the context. This can be problematic when the same attribute contributes to both positive and negative biases within a process, for example, as illustrated in Fig.~\ref{fig:cancer_screening}. \toolname{} addresses this limitation by introducing human-in-the-loop intervention, allowing users to apply context-aware adjustments to the model's decision logic.
While other approaches toward fairness in PBPM exist, such as~\cite{fairness_loss} and~\cite{fairness_gan}, they share similar limitations: they do not involve the user in the decision-making process and remove bias uniformly instead. Moreover, these other methods do not offer a tool implementation with a graphical interface. 

\section{Maturity}
While the technical feasibility of the approach underlying \toolname{} has been demonstrated in our previous work~\cite{Kaeppel2025Fairness}, this paper focuses on a hands-on demonstration using the simulated \textit{cancer screening} event log, as portrayed in Fig.~\ref{fig:cancer_screening}.
Using a simulated event log is necessary, since no real life event log containing sensitive data has been made publicly available.
For the demonstration, we provide an intuitive graphical interface that enables users to interactively inspect and adjust decision trees distilled from predictive models. This allows practitioners without programming experience or familiarity with command-line tools to effectively use \toolname{}.
However, it is important to note that no user study has yet been conducted to evaluate \toolname{} in practice. As such, its usability, effectiveness in supporting fairness interventions, and suitability for real-world deployment remain to be formally assessed.

\mypar{Limitations}
While \toolname{} demonstrates promise, it currently has some restrictions. At present, the tool supports only the PBPM task of the next activity prediction, and only MLPs can be used as the underlying predictive model. 
However, \toolname{}’s architecture is designed with extendability in mind. We plan to broaden the supported prediction task types and model architectures in order to make \toolname{} applicable to a wider range of use cases.
In addition, options for encoding techniques and hyperparameter configuration are currently limited to the most essential settings. To better support model optimization, including hyperparameter tuning, future work should expand these configuration capabilities.
Another key limitation is the absence of built-in metrics for quantifying the fairness of predictive models, meaning the effectiveness of user-driven interventions is currently assessed solely through manual inspection of the distilled decision tree.

\mypar{Future Work}
Besides addressing the aforementioned limitations, we plan to conduct a comprehensive user study to evaluate the usability and effectiveness of \toolname{}. As part of this study, we aim to investigate how fairness metrics can be integrated into the interface to provide intuitive feedback on the consequences of user interventions. We also intend to explore techniques for guiding users through the inspection and modification of decision trees. These techniques are increasingly important, as the complexity of the distilled trees grows with more expressive prediction models and more complex datasets.

\section*{Declaration on the Use of Generative AI}
During the preparation of this work the authors used no generative AI.
%

\bibliography{references}



\end{document}